\documentclass[letterpaper, 10 pt, conference]{ieeeconf}  % Comment this line out
\IEEEoverridecommandlockouts                              % This command is only
\makeatletter
\let\NAT@parse\undefined
\makeatother
\usepackage[numbers,sort&compress]{natbib}

\usepackage{amsmath,amsfonts,amssymb}
\let\labelindent\relax
\usepackage{enumitem}
\usepackage{bm}
\usepackage{prettyref}
\usepackage{graphicx}
\usepackage{mdframed}
\usepackage{wrapfig}
\usepackage{xspace}
\usepackage{subfloat}
\usepackage{subcaption}
\usepackage{cuted}
\usepackage{array,multirow}
\usepackage{booktabs}
\usepackage{mathtools}
\usepackage{bbm}
\usepackage{leftindex}
\usepackage[dvipsnames,table]{xcolor}
\usepackage{cellspace}
\usepackage[T1]{fontenc}
\usepackage{comment}
\usepackage{url}
\usepackage{makecell}
\usepackage{hhline}
\usepackage{cuted}
\usepackage[
    bookmarks=true,
    colorlinks=true,
    backref=page,
    linkcolor=brown,
    citecolor=RoyalPurple!75,
    urlcolor=Maroon
]{hyperref}

\usepackage{algorithm}
\usepackage[noend]{algpseudocode}

\usepackage{stackengine}
\newrefformat{Fig}{Fig.~\ref{#1}}
\newrefformat{fig}{Fig.~\ref{#1}}
\newrefformat{par}{Section~\ref{#1}}
\newrefformat{appen}{Appendix~\ref{#1}}
\newrefformat{sec}{Section~\ref{#1}}
\newrefformat{sub}{Section~\ref{#1}}
\newrefformat{table}{Table~\ref{#1}}
\newrefformat{alg}{Algorithm~\ref{#1}}
\newrefformat{Alg}{Algorithm~\ref{#1}}
\newrefformat{Def}{Definition~\ref{#1}}
\newrefformat{Thm}{Theorem~\ref{#1}}
\newrefformat{Lem}{Lemma~\ref{#1}}
\newrefformat{step}{Step~\ref{#1}}
\newrefformat{ln}{Line~\ref{#1}}
\newrefformat{eq}{Eqn.~\ref{#1}}
\newrefformat{eqn}{Eqn.~\ref{#1}}
\newrefformat{pb}{Problem~\ref{#1}}
\newrefformat{it}{Item~\ref{#1}}
\newrefformat{te}{Term~\ref{#1}}
\def\Eqref Eq:#1:{\eqref{eq:#1}}
\newrefformat{Eq}{Equation~\Eqref#1:}

\newcommand{\setword}[2]{%
  \phantomsection
  #1\def\@currentlabel{\unexpanded{#1}}\label{#2}%
}
\definecolor{Blue}{rgb}{0,0,1}

\newcommand{\lgbnd}{\textcolor{Maroon}{\texttt{\textbf{L-GBND}}}\xspace}
\newcommand{\gbnd}{\texttt{GBND}\xspace}

\title{\LARGE \bf 
Localized Graph-Based Neural Dynamics \\
Models for Terrain Manipulation
}

\author{Chaoqi Liu$^{1}$, Yunzhu Li$^{2}$, and Kris Hauser$^{1}$\\
\href{https://chaoqi-liu.com/scoopbot/}{chaoqi-liu.com/scoopbot}
\thanks{$^{1}$University of Illinois at Urbana-Champaign
        $^{2}$Columbia University.
        This work was partially funded by NSF Grant \#FRR-2409661.
        Corresponding author: {\tt\small chaoqil2@illinois.edu}.
        }
}

\begin{document}

\maketitle

%%%%%%%%%%%%%%%%%%%%%%%%%%%%%%%%%%%%%%%%%%%%%%%%%%%%%%%%%%%%%%%%%%%%%%%%%%%%%%%%
\begin{abstract}
Predictive models can be particularly helpful for robots to effectively manipulate terrains in construction sites and extraterrestrial surfaces. However, terrain state representations become extremely high-dimensional especially to capture fine-resolution details and when depth is unknown or unbounded. This paper introduces \lgbnd, a learning-based approach for terrain dynamics modeling and manipulation, leveraging the Graph-based Neural Dynamics (\gbnd) framework to represent  terrain deformation as motion of a graph of particles. Based on the principle that the moving portion of a terrain is usually localized, our approach builds a large terrain graph (potentially millions of particles) but only identifies a very small active subgraph (hundreds of particles) for predicting the outcomes of robot-terrain interaction. To minimize the size of the active subgraph we introduce a learning-based approach that identifies a small region of interest (RoI) based on the robot's control inputs and the current scene. We also introduce a novel  domain boundary feature encoding that allows \gbnd to perform accurate dynamics prediction in the RoI interior while avoiding particle penetration through RoI boundaries. Our proposed method is both orders of magnitude faster than na\"{i}ve \gbnd and it achieves better overall prediction accuracy. We further evaluated our framework on excavation and shaping tasks on terrain with different granularity. 
\end{abstract}

\section{Introduction}
Terrain manipulation is essential in construction industry and outer space exploration \cite{yifan2023fewshot, lee2024cratergraderautonomousroboticterrain}. Autonomous robots that can create desired structure in the construction site by interpreting sensory data and making decision on where and how to manipulate the terrain, can greatly improve efficiency and be deployed to place where lacks human labor, e.g., other planets. Prediction is a key challenge in terrain manipulation, since robots need to predict outcomes of excavated volume and the movement of terrain shape after scooping, dumping, or pushing material. People have traditionally built approximate models using analytical expressions and geometric calculations to predict excavation outcomes. However, these models are often limited in adapting to other action types and typically require predefined parameters~\cite{reece1964fund_equ_earth}.  Simulation-based methods use terramechanics physics simulators which can handle a wider variety of situations, but these are extremely computationally expensive and still require solving difficult parameter identification problems~\cite{herrmann1992sim-granular, bell2005pb-sim-granular, sommer2022interactive-sim-granular}. Learning-based approaches have also been investigated since early work in autonomous excavation \cite{sing1995synthesis}, but recent attention has been focused on deep learning techniques, such as image-based prediction of granular media manipulation \cite{suh2020surprisingeffectivenesslinearmodels}. A line of recent work has studied the use of Graph-Based Neural Dynamics (GBND), a technique that uses Graph Neural Networks (GNNs) to model system dynamics of a scene represented as set of 3D particles, with each particle representing a small volume of matter \cite{yunzhu2018dpi, deepmind2016in, deepmind2020gns}.  GBND has been applied to various nontrivial deformable object manipulation tasks \cite{shi2023robocook, shi2022robocraft, lin2021softgym}, and can generalize well across different viewpoints, objects and robots poses, shapes and sizes. However, all prior such works have addressed table-top scenarios and objects with bounded volume that can be observed through static cameras.

\begin{figure}[tbp]
    \centering
    \vspace{-10pt}
    \includegraphics[width=\columnwidth]{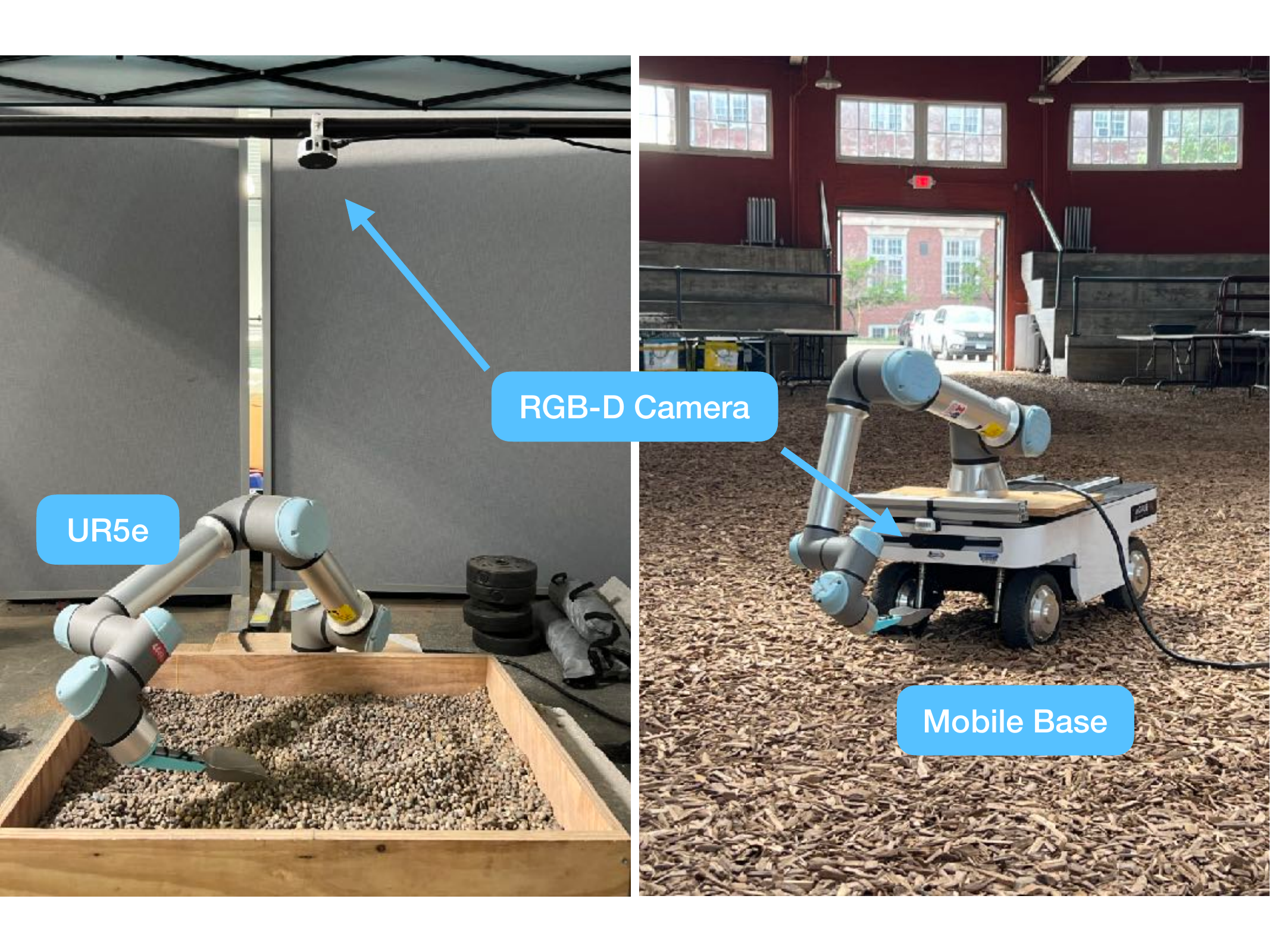}
    \vspace{-20pt}
    \caption{\textbf{Evaluation platforms.} Our terrain manipulation system includes a UR5e robotic arm equipped with an overhead RGB-D camera. We are also extending this system to a mobile scooping platform for in-the-wild studies.}
    \label{fig:mobile_base}
\end{figure}

In this work, we propose a framework, \textcolor{Maroon}{\textit{Localized Graph-Based Neural Dynamics}} (\lgbnd), for modeling deformable terrains at fine resolutions over large spatial scales, such as a construction site in which a mobile excavator uses 3D mapping to generate a terrain representation.  In particular, our method extends the \gbnd to handle spatially large-scale and even unbounded terrains which may be represented by graphs of millions of particles that cannot fit in GPU memory. To predict the deformation of the terrain under a candidate interaction in a computationally efficient manner suitable for planning, our method restricts the attention of dynamics prediction to a dynamically-selected local neighborhood.  We introduce a learning-based 3D region-of-interest (RoI) proposer that identifies a small subset of particles from the large graph that are likely to move.  Assuming particles that are outside the RoI stay static during robot-terrain interaction improves speed by orders of magnitude in addition to higher dynamics prediction accuracy.  \lgbnd simultaneously learns the \gbnd dynamics and the RoI from simulation data, and both are fine-tuned on real data.  Moreover, we introduce a novel boundary feature representation to address the problem that na\"{i}vely applying \gbnd to the identified subgraph leads to erroneous predictions.  In our experiments, we verify both the efficiency and effectiveness of \lgbnd for manipulating terrains of different materials.

Our key innovations include: (1) learning graph dynamics and RoI to minimize \lgbnd memory usage and computation time while retaining high accuracy; and (2) introducing novel node features that enable learning \lgbnd over arbitrary volumetric domains, ensuring that accuracy is preserved at the boundary of RoI, and that translation invariance is maintained.

\begin{figure*}[t]
    \centering
    \includegraphics[width=0.9\textwidth]{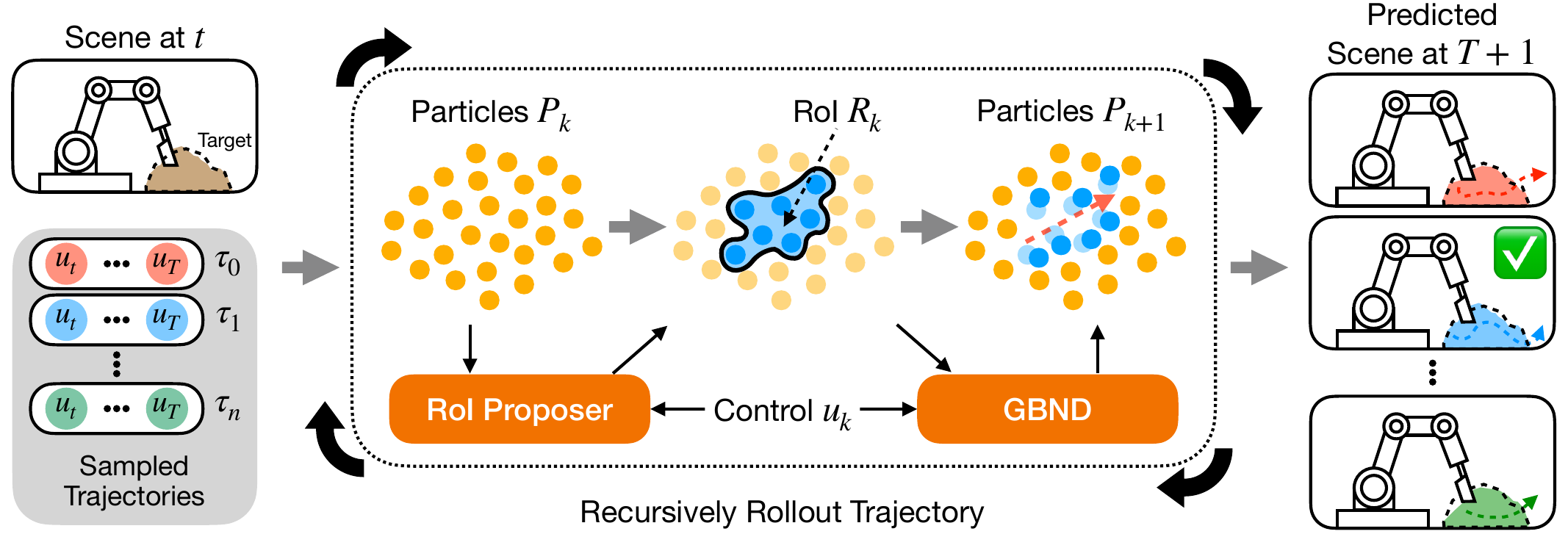}
    \caption{%The proposed \textbf{dynamics} model is depicted in central block: Scene particles together with the control input are feed into the RoI proposer, which select those have potential to move under interaction induced by the control input, followed by rollout via GBND. When coupled with \textbf{planning}, a set of trajectories are sampled and rollout in parallel, the best sampled trajectory under pre-specified metric will be executed.
    \textbf{System overview.} The proposed \textbf{dynamics} model is shown in the central block: scene particles and the control input are fed into the RoI proposer, which selects particles likely to move based on the interaction caused by the control input. This is followed by a rollout using \gbnd. When integrated with \textbf{planning}, multiple trajectories are sampled and rolled out in parallel, and the best trajectory, determined by a pre-defined metric, is executed.}
    \label{fig:pipeline}
    \vspace{-10pt}
\end{figure*}

\section{RELATED WORK AND DISCUSSION}

\textbf{Terrain Modeling and Manipulation.} Effective modeling of terrain movement induced by external forces, and its use in manipulation and locomotion, has been studied for years. Resistive Force Theory (RFT) and 3D-RFT, for example, is generic but reduced-order force models widely used in modeling locomotion in granular media~\cite{yu2024sdm+3drft}. However, RFT lacks the ability to model movements where inertial effects become significant, in addition to assumption of over-simplified geometry \cite{sing1995synthesis}. Other work instead use heightmap as the representation, and regress a latent linear dynamics model \cite{wagner2022bulldozing} for bulldozing. Linear dynamics models lack capability for more complicated interactions~\cite{suh2020surprisingeffectivenesslinearmodels} and discretized heightmaps lack the flexibility of change of resolution, while the optimal resolution varies across tasks and underlying material \cite{yixuan2023dynres}. In addition, particle-based representation provides high-fidelity geometry for tool equipped, focusing more on geometry during interaction, which provides extra precision. Alternatively, model-free methods are utilized widely \cite{xu2024tactilebasedobjectretrievalgranular}, but these approaches often grapple with high sample complexity and exhibit a task-specific nature.

\textbf{Dynamics Model Learning.} Physics-based models have shown effectiveness in a wide range of robotic manipulation tasks \cite{pang2023global, yu2016more, hogan2020feedback}, but accurate model building requires full state estimation. To address this issue, recent advances in machine learning are applied to learn the system dynamics models directly from sensory data~\cite{nagabandi2020deep, hu2024learninggranularmediaavalanche, wang2025learningrealworldactionvideodynamics}. The learned dynamics model generally demonstrates superior effectiveness relative to its model-free counterparts~\cite{liu2023modelbasedcontrolsparseneural}. Graph-based scene representation processed by GNNs have showcased effectiveness in dynamics modeling due to the spatial relational inductive bias injected \cite{battaglia2018relational, pfaff2020learning, yunzhu2018dpi, deepmind2016in, deepmind2020gns}. Prior work explored a wide range of applications of graph-based dynamics model, including but not limited to rigid bodies \cite{yunzhu2018dpi, deepmind2016in}, elastic-plastic materials \cite{shi2022robocraft, shi2023robocook}, fluids \cite{yunzhu2018dpi, deepmind2020gns}, and granular materials \cite{yixuan2023dynres, deepmind2020gns, Tuomainen_2022}. However, the vast majority of prior work model systems that are either confined to a 3D box or atop a 2D plane, and our work demonstrates an approach to extend graph-based approaches to terrains of unbounded breadth and depth using regions of interest and RoI boundary features.

\textbf{Model-Free Policy Learning.} While model-free methods such as reinforcement learning (RL) and behavior cloning (BC) have achieved notable success in robotic manipulation~\cite{chi2024dp, du_learning_2023, jannerPlanningDiffusionFlexible2022a, carvalhoMotionPlanningDiffusion2023, høeg2025hybriddiffusionsimultaneoussymbolic, liu2025flexiblemultitasklearningfactorized, chen2025multimodalmanipulationmultimodalpolicy}, they typically suffer from limited generalization, particularly in environments involving complex, high-dimensional, and long-horizon interactions like terrain deformation. These approaches require extensive task-specific data collection and tuning, often making them impractical for settings involving diverse or previously unseen terrain geometries. In the context of deformable terrain manipulation, the inability of model-free policies to extrapolate beyond the training distribution presents a fundamental limitation~\cite{NEURIPS2020_1e7875cf}. While such methods can learn reactive strategies within the training distribution, they lack an explicit mechanism for generalizing to new goals or terrain states. In contrast, dynamics model-based approaches retain predictive structure: given a learned forward model, planning can be performed zero-shot for unseen terrain states by simulating the effects of actions. This capability enables flexible and task-agnostic generalization, which is essential for terrain-aware manipulation. Because of these limitations, we do not consider model-free policies as baselines in this work. Our focus is exclusively on dynamics model learning, as it enables both general-purpose planning and scalability across diverse terrain interaction scenarios.

\begin{figure}[t]
    \centering
    \includegraphics[width=.95\columnwidth]{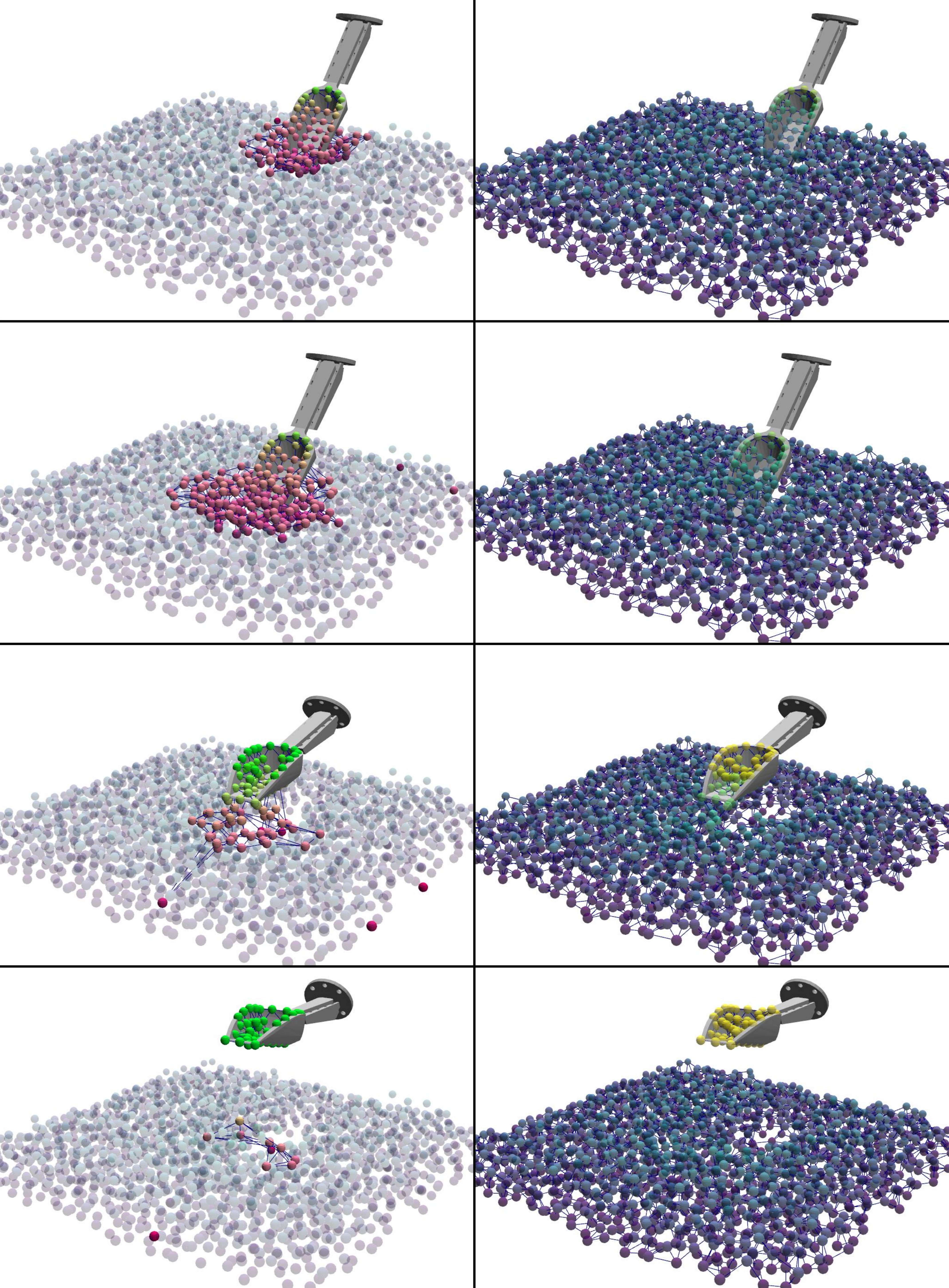}
    \caption{\textbf{Visualization of a typical scooping simulation.} Our method (left) only predicts the dynamics of the highlighted particles in the predicted RoI.  Predictions are computed for an order of magnitude fewer particles as the full graph (right) while retaining similar accuracy.}
    \label{fig:scoop_filmstrip}
\end{figure}
\section{Localized Graph-Based Neural Dynamics}
\label{sec:method}

We present a framework for efficient and accurate terrain dynamics prediction under robot interaction, based on \textit{localized} \gbnd (\lgbnd) models. Our goal is to learn a computationally-efficient model to predict the shape of a terrain after robot-terrain interaction, such that \lgbnd can be used in terrain manipulation planning. Specifically, letting $x_t$ denote the state of the terrain, we model $\hat{x}_{t+1} = f(x_t, u_t)$ where $u_t$ is the robot control and $t$ is the time index. We learn fine-grained prediction where the time step is 0.1\,sec and $u_t$ indicates a small robot end-effector displacement. The dynamics model is trained on data recorded during three types of actions (scooping, pushing, and dumping), and we use a unified dynamics model for all actions. Pretraining is performed from simulation data and then the model is fine-tuned on real data.

We choose \lgbnd as our terrain model, which is a particle-based scene representation that provides flexibility in inter-particle interactions, an inductive bias toward locality in interactions, mass conservation, and preserves fine details of deforming solids and fluids~\cite{yunzhu2018dpi, deepmind2020gns, deepmind2016in}. To manage the high-dimensional state space of the terrain, we assume that only a small region of interest (RoI) $R_t(u_t) \subset \mathbb{R}^3$ of the volume is affected by the interaction. Letting $p_t^{(i)}$ denote the $i^\text{th}$ particle in the terrain at time $t$, its future state $\hat{p}_{t+1}^{(i)} = p_t^{(i)}$ for any point $p_t^{(i)} \notin R_t(u_t)$. Our method simultaneously learns $f$ and $R_t$. The RoI proposer is formulated as a super-level set of a learned function $g(p,u_t)$ such that $g(p,u_t) \geq 0$ if and only if $p \in R_t(u_t)$. Using the RoI, we only predict the motion of a subgraph of the particle graph. The proposed framework is shown in Fig.~\ref{fig:pipeline}.

\subsection{Graph-Based Scene Representation}

We adopt a particle-based representation of the environment, in which both terrain and robot end-effectors are modeled as sets of particles. This representation is particularly suitable for learning physical dynamics in deformable and granular media, as each particle implicitly represents a small volume of matter. Terrain particles $P_t = \{ p_t^{(i)}\}_{i=1}^{N_p}$ are instantiated from the observed terrain surface map, volumetrically down to a sufficiently large depth. Tool particles $T_t = \{ p_t^{(i)} \}_{i=N_p+1}^{N_p+N_t}$ are sampled on the robot’s end-effector. For each particle, we connect it to the $k$ closest particles that meet a distance threshold using a nearest-neighbor algorithm, resulting in a sparse graph.

Follow prior work~\cite{deepmind2016in, yunzhu2018dpi, deepmind2020gns}, we can define nodes in the graph as
\begin{align}
v_t^{(i)} \triangleq \left[ \mathsf{v}_{t-H:t-1}^{(i)}, u_t^{(i)}, c^{(i)} \right]^\top, \label{equ:naive_node_features}
\end{align}
where $\mathsf{v}_{t-H:t-1}^{(i)}$ is particle $p^{(i)}$’s velocity over the last $H$ time steps, $u_t^{(i)}$ is the impulse on the tool particle directly caused by robot action (displacement of the end-effector, hence, naturally zeros for terrain particles), and $c^{(i)}$ is the particle class (terrain or tool) embedding. The edge features for an edge $e_t^{(k)}$ connecting particles $i$ and $j$ are
\begin{align}
e_t^{(k)} \triangleq \left[ p_{t-H:t}^{(i)} - p_{t-H:t}^{(j)}, c^{(i)}, c^{(j)} \right]^\top.
\end{align}

\subsection{Graph-Based Neural Dynamics}

The dynamics of the particle system are modeled using a Graph Neural Network (GNN), which represents the environment state as a directed graph $x_t \equiv G_t(V_t, E_t)$ at each time step, where $V_t \triangleq \{ v_t^{(i)} \}_{i=1}^{|V_t|}$ is the set of nodes, and $E_t \triangleq \{ e_t^{(k)} \}_{k=1}^{|E_t|}$ is the set of edges, with each edge $e_t^{(k)}$ denoting a potential direct interaction from $v_t^{(i)}$ to $v_t^{(j)}$.  Indirect interactions between particles are modeled in this framework using a message passing algorithm.

The dynamics are defined as follows. Each node and edge are first encoded into a latent representation,
\begin{align}
    z_0[v_t^{(i)}] = f_{V}^{\text{enc}}( v_t^{(i)} ), && z_0[e_t^{(k)}] = f_{E}^{\text{enc}} ( e_t^{(k)}).
\end{align}
Then, message propagation is run for $L$ iterations to propagate, aggregate, and update latent representations across the graph. Mathematically, for $l \in \{ 0, 1, ..., L-1 \}$, 
\begin{equation}
    z_{l+1}[e_t^{(k)}] = f_{E}^{\text{prop}} ( z_{l}[e_t^{(k)}], z_{l}^{i}, z_{l}^{j} )
\end{equation}
for an edge $e_t^{(k)} \triangleq (i, j)$, and messages along edges to nodes are propagated and aggregated via 
\begin{equation}
z_{l+1}[v_t^{(i)}] = f_{V}^{\text{prop}} \Big{(} z_{l}[v_t^{(i)}],  \sum_{e \in E^+(v_t^{(i)})} z_l[e]  \Big{)}.
\end{equation}
Finally, a node decoder $f_{V}^{\text{dec}}$ predicts  displacement of the particle, $\Delta\hat{p}_t^{(i)} = f_{V}^{\text{dec}}( z_L[v_t^{(i)}] )$.  The ultimate dynamics prediction is given by $\hat{p}_{t+1}^{(i)} = p_t^{(i)} + \Delta\hat{p}_t^{(i)}$.
\gbnd learns encoder, decoder, and propagation networks $f_V^{\text{enc}}$, $f_E^{\text{enc}}$, $f_V^{\text{dec}}$, $f_V^{\text{prop}}$, $f_E^{\text{prop}}$ via backpropagation through message passing.

\begin{algorithm}[t]
\caption{\small L-GNBD}
\label{algo:gnn+cnn}
\begin{algorithmic}
\Procedure{Propose}{$ \{ p_{t-H:t}^{(i)} \}_{i=1}^N, u_t $}
    \State{$M_{t-H:t} \gets $ \textsc{Project2D}( $\{ p_{t-H:t}^{(i)} \}_{i=1}^N$ )}
    \State{Predict $\tilde{g}$ based on $M_{t-H:t}, u_t$  \Comment{$R_t$ in Sec. \ref{sec:roi}}}
    \For{$i \in [N]$}
        \If{$p_t^{(i)}.z \geq \tilde{g}(p_t^{(i)}.x \,,\, p_t^{(i)}.y)$}
            \State{$\emph{indices} \gets \emph{indices} \cup \{i\}$}
        \EndIf
    \EndFor
    \State{\textbf{return} $\emph{indices}$ }
\EndProcedure
\\
\Procedure{Step}{ $ x_{t-H:t}, u_t $}
    \State{Instantiate scene particles $\{ \leftindex^Wp_{t-H:t}^{(i)} \}_{i=1}^N$}
    \State{Align coordinate frame $L$ with scoop-tip}
    \State{Localize particles $ \leftindex^Lp_{t-H:t}^{(i)} \gets \leftindex^LT^W  \leftindex^Wp_{t-H:t}^{(i)}$}
    \State{\emph{indices} $\gets$ \textsc{Propose}( $ \{  \leftindex^Lp_{t-H:t}^{(i)} \}_{i=1}^N, u_t $ )}
    \State{Rollout dynamics $\forall i,$\[
        {}^L\hat{p}_{t+1}^{(i)} = 
        \begin{cases}
            f_\theta( \leftindex^Lp_{t-H:t}^{(i)}, u_t^{(i)} ) &\mbox{if $i \in$ \emph{indices}} \\
            ^Lp_t^{(i)} &\mbox{otherwise}
        \end{cases}
    \]}
    \State{Restore particles $ \leftindex^Wp_{t-H:t}^{(i)} \gets \leftindex^WT^L  \leftindex^Lp_{t-H:t}^{(i)}$}
    \State{\textbf{return} $\{ {}^W\hat{p}_{t+1}^{(i)} \}_{i=1}^N$}
\EndProcedure
\end{algorithmic}
\end{algorithm}

\begin{algorithm}[t]
\caption{\small MPPI Planning with L-GBND}
\label{algo:plan_w_gbnd}
\begin{algorithmic}
\Require score $S$, scene $x_t$, goal description $\mathsf{G}$, motion parameter space $\Omega$, batch size $b$, time budget $\mathsf{B}$
\State{$u^*_{0:T}\gets nil$}
\While{time used < $\mathsf{B}$  }
    \State{Sample a batch of trajectory parameters $\omega_1,\ldots,\omega_b$}
    \State{Convert batch of $\omega_i$ to control sequences $u_{i,0:T_i}$, $\forall i$ }
    \State{Predict batch $\hat{x}_{i,t+T_i} \gets $ Rollout($x_t,u_{i,0:T_i}$), $\forall i$}
    \State{Evaluate scores $S(\hat{x}_{i,t+T_i},\mathsf{G})$, $\forall i$}
    \State{Update best control sequence $u^*_{0:T}$ }
\EndWhile
\State{\textbf{return} $u^*_{0:T}$}
\end{algorithmic}
\end{algorithm}

\subsection{Adaptive Fine-Grained Region of Interest Identification}
\label{sec:roi}

\gbnd’s computational and memory costs scale with the number of particles and edges; therefore, constraining the graph size is critical for efficiency. We learn an implicit function $g(p, u_t)$ that defines the region of interest (RoI) $R_t(u_t) = \{ p \in \mathbb{R}^3 \mid g(p, u_t) \geq 0 \}$.

We first extract the planar tool position and yaw from $u_t$ and define a 2D reference frame $T_\text{tool}$. We then predict a tool-centered 2D height function $\tilde{g}(x, y)$ such that $g(p, u_t) = p_z - \tilde{g}(T_\text{tool}^{-1}(p_x,p_y))$. In other words, points whose $z$ exceeds $\tilde{g}(x, y)$ (in the tool-centered frame) are included in the RoI.

Our approach uses a terrain heightmap and instantiates $\tilde{g}$ as a Convolutional Neural Network (CNN) architecture. A heightmap with grid resolution 2.5\,cm is centered about the tool and cropped to a 0.4\,m $\times$ 0.4\,m region. At each time step, particles in the bounding box are squeezed into a 32$\times$32 heightmap, with each pixel concatenated with a 16D latent representation of the current control input. Our CNN is a U-Net-like~\cite{ronneberger2015unet} architecture with three levels of pooling, nonlinearity, and upsampling steps with skip connections. The output is a 16$\times$16 depth map representation of $\tilde{g}$.

Training uses simulation data. Moving particles are identified using a velocity threshold, and connected component filtering is applied to remove noise. The target heightmap is defined by combining the upper heightmap of non-moving particles (inflated to produce a bowl shape) with the bottom-up heightmap of moving particles. An $L_1$ loss is applied to non-empty cells, while a hinge loss at 0.5\,m is used for empty cells. The pseudocode of \lgbnd with scale reduction is presented in Algo.~\ref{algo:gnn+cnn}, and a typical simulation rollout with adaptive RoI scale reduction is presented in Fig.~\ref{fig:scoop_filmstrip}.

\begin{table}[t]
\centering
% \scriptsize
\footnotesize
\begin{tabular}{l|ccc}
\toprule
                    & \gbnd & \gbnd w/ $z$ features & \lgbnd \\
\midrule
In Distribution     & 1.85     & 1.07       & 1.09              \\
Out of Distribution & 1.85     & 2.85       & 1.09              \\
\bottomrule
\end{tabular}
\caption{Comparison of node feature encodings for boundaries. \gbnd fails to generalize, and adding absolute height ($z$) overfits to training terrain elevation. \lgbnd with normal features performs consistently across distributions.}
\label{tab:node_features}
\end{table}

% \begin{figure}[t]
%     \centering
%     \includegraphics[width=\columnwidth]{images/node_features.pdf}
%     \vspace{-15pt}
%     \caption{Comparing different methods for encoding RoI boundaries in node features. The baseline (no encoding) does not perform well, whereas encoding absolute height ($z$ features) leads to implicitly learning a ``floor effect.'' This becomes detrimental when the validation set has different heights from the training set. In contrast, our proposed normal features perform well in all cases.}
%     \label{fig:node_features}
%     \vspace{-15pt}
% \end{figure}

\subsection{Boundary-Aware Node Features}

\gbnd is best suited for tasks with similar training / test distributions, but our setting introduces the complication of unboundedness and varying regions of interest.  As an example, past works using \gbnd for fluid and pile manipulation have used a fixed domain of particles, such that the network could {\em implicitly} learn the boundary conditions of a floor and walls~\cite{shi2022robocraft, shi2023robocook, ai2024robopack, yixuan2023dynres}. In our case, ``boundary conditions'' are not strictly-enforced boundaries, but are rather just more material that is unaffected by the current interactions due to energy dissipation.

In effect, particles adjacent to the RoI boundary should ``feel'' resistance as though the boundary were fixed. With naive node feature encoding (Equ.~\ref{equ:naive_node_features}) as used in prior work, these particles will simply pass through the boundary as if there was nothing there. Alternatively, we experimented with adding ``wall particles'', a boundary set of nodes, directly connected to nodes in the RoI, whose latent features are frozen. Not only does this increase memory usage due to massive wall particles, but in our experiments, inner particles were able to penetrate gaps in the sampled wall particles which introduces unnatural artifacts.

To encode appropriate boundary behavior, our approach incorporates an estimated geometric normal of particles within RoI $R_t$~\cite{tombari2010pcd_normal}. In the interior, the normal information is not likely to help much, but normals are likely to be strongly aligned near the boundary. This allows our network to encode the effect that nodes near the RoI boundary should ``feel'' resistance in the direction of the normal. This encoding also has the favorable effect of translational invariance, which is that for the overall dynamics model $f$,
\begin{align}
    \forall \Delta p \in \mathbb{R}^3, f(p_t^{(i)} + \Delta p, u_t) = f(p_t^{(i)}, u_t) + \Delta p.
\end{align}
This ensures that the model's predictions are independent of the absolute positions of particles. 

Experiments in Table \ref{tab:node_features} present validation losses of learned \gbnd models for different sets of node features.  As baselines we compare standard \gbnd encoding as well as augmenting with particle height ({\em $z$ features}). In the {\em in distribution} case, the training and validation sets used terrains whose particles were sampled in a fixed $z$ range of [0\,cm, 5\,cm].  In the {\em out of distribution} case, the validation particle sets were randomly shifted in the $z$ axis by randomly sampled offsets from -100 to 100\,m.  Encoding the $z$ value significantly improves loss in the in-distribution case, demonstrating that the network is implicitly learning a ``floor effect''. This implicit relationship is detrimental in the out-of-distribution case.  In contrast, our \lgbnd with proposed normal encoding leads to better generalization than the baseline in both regimes.

\subsection{Sim-to-Real Transfer}
\label{sec:sim2real}

To transfer the pretrained \lgbnd to the real world, we allow the robot to autonomously collect its own interaction data through self-play. Real-world trajectories are monitored using an overhead RGB-D camera, from which scene point clouds are reconstructed. Robot proprioception is used to remove robot particles and isolate in-tool particles by re-sampling within the tool's convex hull. Since real-world data lacks explicit particle correspondences across time, we follow standard practices in the \gbnd literature and fine-tune the model using Chamfer Distance~\cite{fan2017point/cd} and Earth Mover's Distance~\cite{rubner2000emd} between predicted and observed point clouds. This enables effective learning without requiring explicit tracking of individual particles across frames.

\subsection{Terrain Manipulation Planning with Localized GBND}

We use our \lgbnd in a planner for terrain manipulation tasks. Our planner selects among parameterized scooping, pushing, and dumping trajectories to minimize the difference between the predicted terrain shape and a target heightmap. For scooping and pushing, we use the penetrate-drag-scoop (PDS) trajectory used in excavation~\cite{yifan2023fewshot, lu2021excavation, sing1995synthesis}, which maintains a 0 roll angle. This yields a 9D motion parameter space. Dumping uses a 6D parameterization (dumping location, yaw angle, pre- and post-dump pitch angles). Scooping and pushing are only allowed if the scoop is empty. The planner is based on Model Predictive Path Integral (MPPI) control~\cite{williams2016mppi}. A batch of trajectories is sampled in each iteration. The corresponding control sequences are rolled out with the dynamics model, and the outcomes are scored based on proximity to the goal shape. The best trajectory is selected and executed, and the terrain state is updated. Pseudocode is shown in Algo.~\ref{algo:plan_w_gbnd}.

\section{EXPERIMENTS AND RESULTS}

The computational performance and accuracy of the proposed method were thoroughly evaluated and contrasted against simpler baselines. Furthermore, the capability of our method was assessed through its application to extended-range excavation tasks. All experimental evaluations were conducted using an Intel i9-13900K CPU and a single NVIDIA GeForce RTX 4080 GPU.

\subsection{Implementation Details}

The architectures of \gbnd are MLPs, trained over 200,000 backpropagation steps, unless stated otherwise. For data collection during pre-training, the SAPIEN simulator~\cite{xiang2020SAPIEN} was utilized. Attempts to employ Nvidia Flex \cite{macklin2014flex} were made, but it was abandoned due to the increasing prevalence of unrealistic behaviors as the particle count escalated. In the terrain generation phase, the diamond-square algorithm \cite{diamondsquarealgo} was employed to produce landscapes comprised of approximately 3,000 miniature cubes. Each cube was then populated with 5 to 10 particles, and further reduction of terrain particle sets was achieved through farthest-point sampling \cite{qi2017pointnet++}. 

Approximately 1,000 robot-terrain interaction trajectories, each uniquely parameterized, were simulated. Throughout these simulations, the downsampled terrain particles were meticulously tracked to facilitate the learning of the dynamics model. Training on this simulation data involved regressing the motion trajectories of the particles using the Mean-Square Error (MSE) criterion, aided by the perfect association of scene particles. Concurrently, the CNN proposer was trained as detailed in Section~\ref{sec:roi}. The model's forward simulation steps were set at 0.1 seconds.

\subsection{Particle-based v.s. 2D Heightmap Representations}

\begin{figure}[t]
    \centering
    \includegraphics[width=\linewidth]{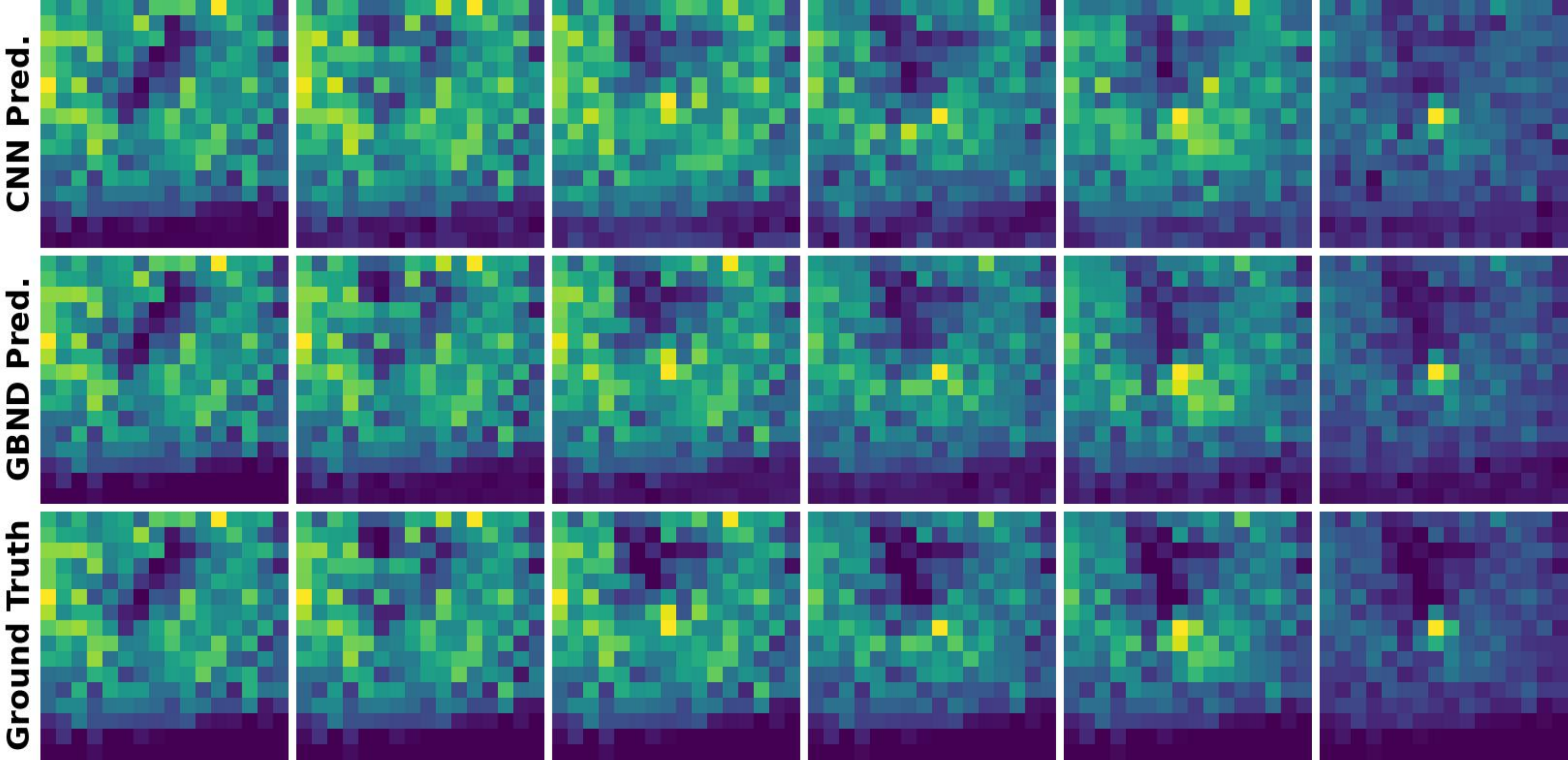}
    \caption{Rollouts (left to right) from a learned 2D CNN heightmap-based dynamics model compared to \lgbnd and ground truth. In the CNN, fine-grained features are lost due to the compressed representation and smoothing bias of CNNs. In contrast, \lgbnd preserves volumetric structure and local interactions for 37\% lower prediction error on a test dataset.}
    \label{fig:2d_heightmap_dynamics}
\end{figure}

We compare \lgbnd against a baseline that represents terrain using a 2D heightmap and learns dynamics directly in heightmap space. Both models are trained on the same dataset using the same control inputs. For consistency and implementation efficiency, the 2D baseline reuses the architecture of our RoI proposing CNN: a U-Net-style convolutional network that takes as input a stack of past terrain heightmaps and current control input, and autoregressively predicts the future heightmap. To ensure a fair comparison, we project the output of the particle-based model onto a heightmap before computing evaluation metrics. We evaluate prediction accuracy using the $L_1$ loss over the heightmap grid. Quantitatively, \lgbnd achieves approximately 37\% lower prediction error compared to the 2D heightmap model. Fig.~\ref{fig:2d_heightmap_dynamics} presents qualitative results comparing heightmap-based dynamics prediction and ground-truth terrain heightmaps.

This comparison highlights the critical role of representation in deformable terrain dynamics modeling. The 3D particle-based representation used in \gbnd captures volumetric structure and encodes fine-grained geometry, providing strong inductive biases toward spatial locality, directional interaction patterns, and mass conservation. These properties enable the model to represent contact-rich interactions such as scooping, penetration, and material buildup along tool edges—phenomena that are difficult to express in 2D. In contrast, 2D heightmap representations collapse the terrain into a single surface, discarding sub-surface interactions and occluding contact geometry. Moreover, convolutional architectures tend to produce overly smoothed predictions, which dampens sharp discontinuities and local terrain features. These limitations manifest even more clearly when directly predicting future terrain shape.

\subsection{\lgbnd's Efficiency and Effectiveness}

\begin{figure}[tbp]
    % \raggedright
    \hspace{-10pt}
    \centering    
    \includegraphics[width=.85\columnwidth]{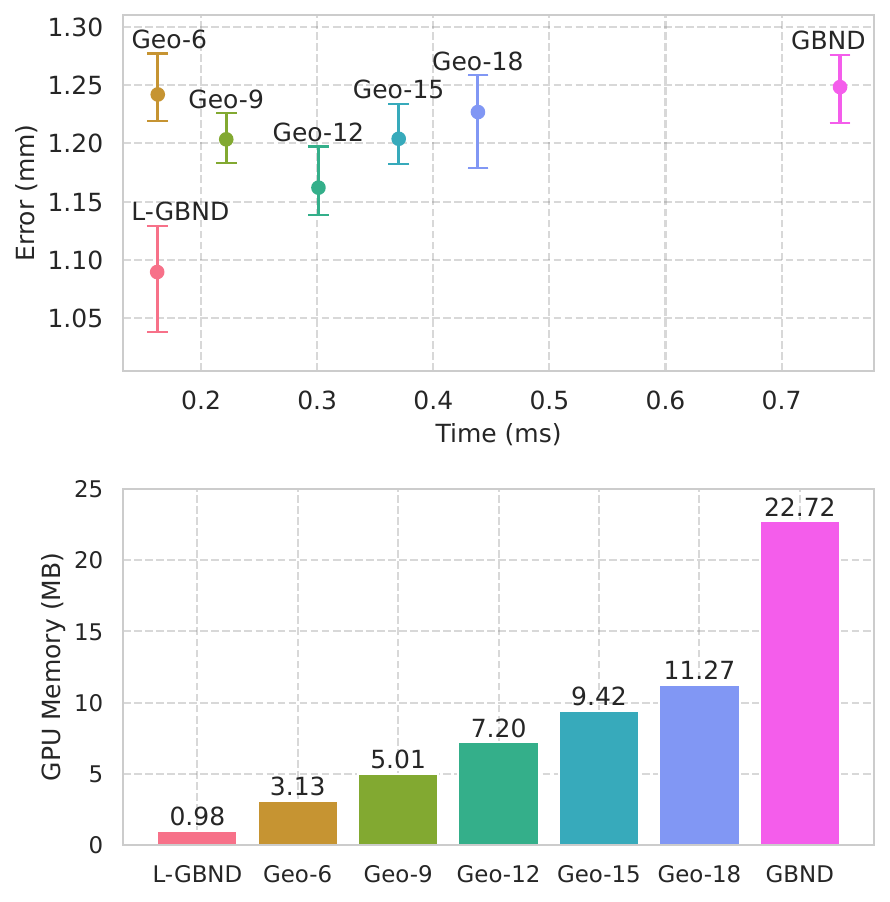}
    \caption{Comparing our \lgbnd against \gbnd and geometric region proposers with different size (Geo-X). Our method demonstrates significant advantages in speed and GPU memory. $\approx$ 3,000 particles and batch size 256 are used in these experiments, per sample measurements are reported (i.e., divided by 256). Colors and labels are shared by both figures.
    }
    \label{fig:efficiency_effectiveness_tradeoff}
\end{figure}

The computational performance and accuracy of our method are compared against two baselines. The first baseline utilizes all particles within a large-scale terrain, while the second employs a na\"{i}ve geometric Region of Interest (RoI), specifically, a cylinder centered at the scoop tip with a predetermined radius. To maintain computational feasibility with the full-terrain baseline, terrains are restricted to 3,000 particles — significantly fewer than what might be encountered by an excavator operating on an large worksite. Figure~\ref{fig:efficiency_effectiveness_tradeoff} presents the results on a terrain measuring 50\,cm $\times$ 50\,cm $\times$ 5\,cm, with a particle density of 0.5 particles per cm$^3$. Here, `Geo-X' represents the geometric RoI with a cylinder diameter of X\,cm. Each RoI-based approach is developed using the methodology described in Section~\ref{sec:roi} alongside our novel node boundary features.

In terms of per-sample processing time and GPU memory utilization, our method matches the speed of the smallest geometric RoI and requires even less GPU memory, while delivering superior accuracy. It is important to note that enhanced RoI prediction significantly improves the accuracy of dynamics predictions. This is because the \gbnd is prone to inaccuracies, such as predicting minimal movements of static particles. An RoI introduces a strong inductive bias that is absent in \gbnd, leading to jitter among distant particles. For smaller values of X, Geo-X exhibits large errors due to the exclusion of moving particles. Conversely, as X increases, the inclusion of numerous static particles leads to performance degradation due to \gbnd jitter.

Furthermore, since the planning speed and quality are heavily influenced by the batch size, memory efficiency becomes a critical objective for our dynamics model. Table~\ref{tab:prediction_speed} illustrates that our method reaches optimal performance at a batch size of $b=256$, achieving an average processing time of less than 0.2\,ms per sample. Under these conditions, the planner is capable of rolling out hundreds of trajectories across numerous time steps in seconds.  We also compare against the recently updated GPU-based simulator in SAPIEN 3.0~\cite{xiang2020SAPIEN} in scenes with 3,000 cube-shaped particles, without a simulated robot.  Our method is an order of magnitude faster and can handle larger batch sizes within available GPU memory. As a result of superior efficiency, \lgbnd leads to better planning performance in terrain shaping tasks as shown in Table~\ref{tab:planning_performance_wrt_batchsize}, where we analyze the scaling of planning performance with respect to number of candidate trajectories being simulated.

\begin{table}[t]
\centering
\scriptsize
\begin{tabular}{@{}l|cccccc@{}}
\toprule
Batch & \lgbnd  & Geo-6 & Geo-12         & Geo-18         & \gbnd    & SAPIEN         \\ \midrule
1     & 4.624 & 3.543 & 4.459          & 5.157          & 5.443 & 3.309 (CPU)    \\
2     & 2.454 & 1.913 & 2.280          & 2.616          & 2.775 & 2.104          \\
4     & 1.400 & 1.036 & 1.250          & 1.463          & 1.642 & 1.627          \\
8     & 0.860 & 0.586 & 0.755          & 0.931          & 1.097 & \textbf{1.563} \\
16    & 0.522 & 0.366 & 0.501          & 0.650          & 0.872 & 1.713          \\
32    & 0.337 & 0.242 & 0.364          & 0.520          & 0.885 & 1.991          \\
64    & 0.232 & 0.181 & 0.332          & 0.498          & 0.809 & 2.211          \\
128   & 0.179 & 0.169 & 0.318          & 0.456          & 0.774 & 1.755          \\
{\color[HTML]{3166FF} 256} &
  {\color[HTML]{3166FF} \textbf{0.162}} &
  {\color[HTML]{3166FF} \textbf{0.163}} &
  {\color[HTML]{3166FF} 0.301} &
  {\color[HTML]{3166FF} 0.438} &
  {\color[HTML]{3166FF} \textbf{0.750}} &
  {\color[HTML]{3166FF} OoM} \\
512   & 0.181 & 0.181 & \textbf{0.293} & \textbf{0.431} & OoM   & OoM            \\
1024  & 0.198 & 0.198 & 0.294          & OoM            & OoM   & OoM            \\ \bottomrule
\end{tabular}
\caption{Average prediction time (ms) per sample as the batch size varies, showing that 256 is an optimal batch size for our framework. For SAPIEN, 3000 cubes are simulated.}
\label{tab:prediction_speed}
\end{table}

\begin{table}[t]
\centering
\scriptsize
\setlength{\tabcolsep}{2.0pt}
\begin{tabular}{@{}l|lllllllllll@{}}
\toprule
Batch                             & 1     & 2     & 4     & 8     & 16    & 32   & 64   & 128  & \textbf{256}           & 512  & 1024 \\ \midrule
\# rollouts & 216   & 407   & 714   & 1163  & 1916  & 2967 & 4310 & 5587 & \textbf{6173} & 5525 & 5051 \\
$L_1 (10^{-3})$ & 15.32 & 14.53 & 12.68 & 11.37 & 10.91 & 7.22 & 5.72 & 3.93 & \textbf{3.42} & 4.11 & 4.31 \\ \bottomrule
\end{tabular}
\caption{For given time budget (1\,sec), more samples being simulated with \lgbnd in MPPI planning leads to better planning performance.}
\label{tab:planning_performance_wrt_batchsize}
\end{table}

\subsection{Real World Terrain Manipulation}
\label{sec:exp_plan_w_gbnd}

\begin{figure}[t]
    \centering
    \begin{subfigure}{.4\columnwidth}
        \centering
        \includegraphics[width=\columnwidth]{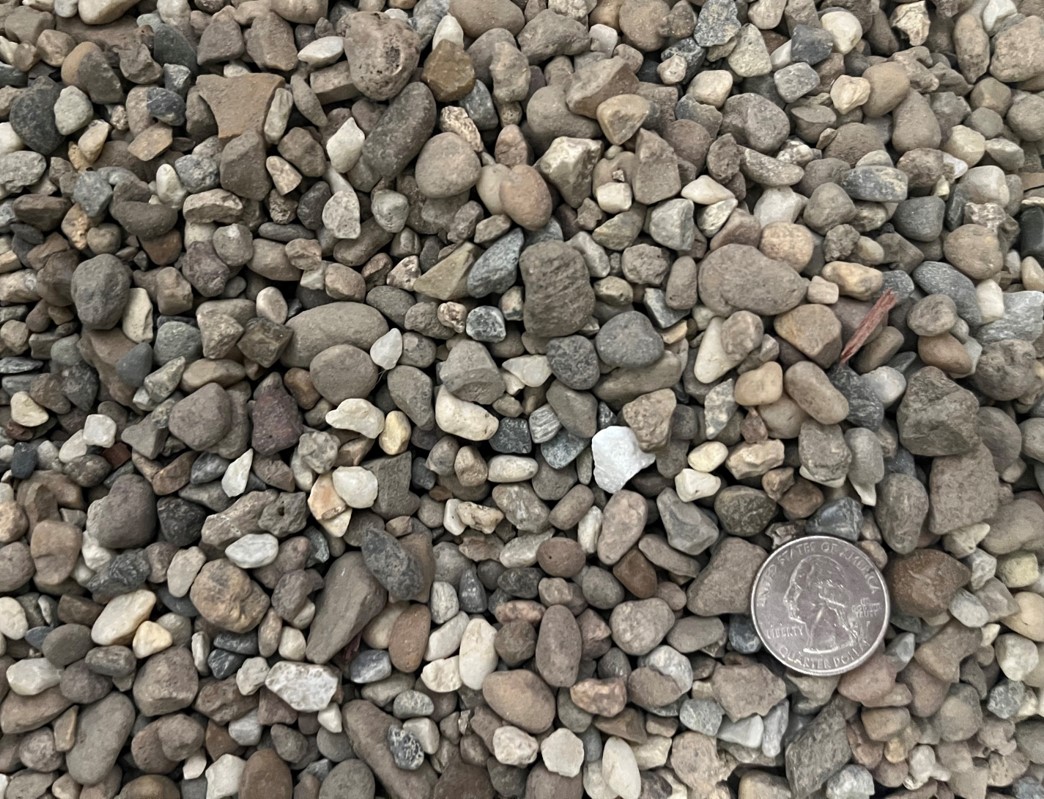}
        \vspace{-15pt}
        \caption{Pebbles 0.8 - 1.0\,cm}
    \end{subfigure}
    % \hfill
    \hspace{10pt}
    \begin{subfigure}{.4\columnwidth}
        \centering
        \includegraphics[width=\columnwidth]{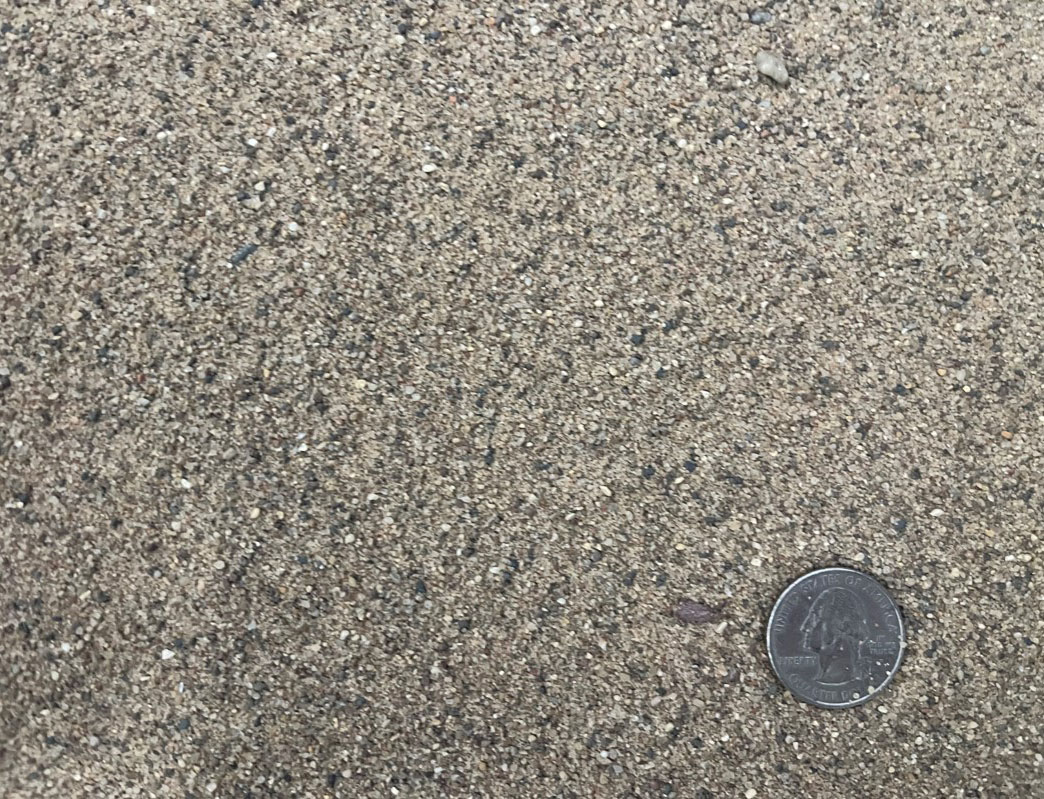}
        \vspace{-15pt}
        \caption{Sand $\ll$ 1\,mm}
    \end{subfigure}
    \vspace{-3pt}
    \caption{Materials used in real world experiments, U.S. quarter coin on the bottom right for scale. 
    % Figure borrowed from \cite{yifan2023fewshot}.
    }
    \label{fig:materials}
    \vspace{-5pt}
\end{figure}

Real-world experiments were conducted using the platform shown in Fig.~\ref{fig:mobile_base}, left, on two different materials as depicted in Fig. \ref{fig:materials}: (1) pebbles with an approximate grain size of 0.8 - 1.0\,cm, and (2) play sand with a grain size significantly less than 1\,mm. To ensure experimental repeatability, the terrain was contained within a box measuring 0.9\,m $\times$ 0.6\,m $\times$ 0.2\,m, though the methodologies developed are equally applicable to natural terrains.

To evaluate real-world transfer, we fine-tune the pretrained model on 100 real-world interaction trajectories collected via autonomous self-play and overhead RGB-D sensing. Following the procedure described in Section~\ref{sec:sim2real}, we compute supervision signals using Chamfer and Earth Mover’s Distance between predicted and observed point clouds. This fine-tuning enables the model to adapt effectively to real-world sensor data and contact dynamics.

\begin{figure}[t]
    \centering
    % \vspace{-10pt}
    \hspace{-10pt}
    \includegraphics[width=.75\columnwidth]{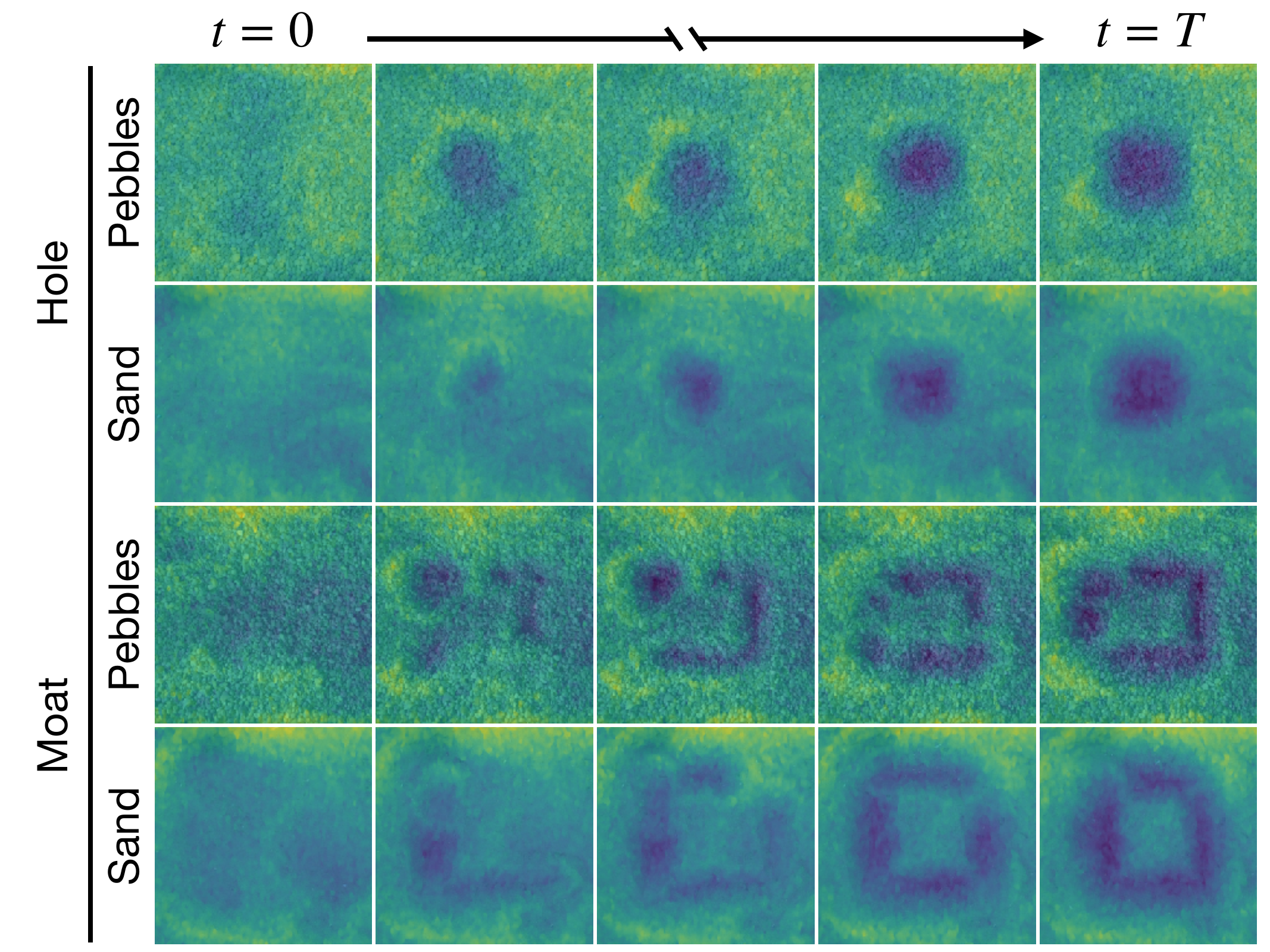}
    \caption{Sequences of heightmaps recorded during real-world manipulation.}
    \label{fig:task_filmstrip}
    \vspace{-10pt}
\end{figure}

The experimental setup focused on two specific terrain shaping tasks: (1) creating a hole, and (2) forming a rectangular moat. The desired terrain configurations were represented as 2D heightmaps, with the $L_1$ loss between the current scene heightmap and the target heightmap serving as the performance metric.  These target shapes have sharp discontinuities and cannot be physically realized due to the material's settling behavior (angle of repose is approximately 25-40$^\circ$). For each manipulation task, our planner evaluated 1500 trajectories with an average horizon of 30 steps (equivalent to 3 seconds), requiring 20-30 seconds of planning time.  Qualitative results are shown in Fig. \ref{fig:task_filmstrip}, demonstrating successful terrain shaping.

\begin{figure}[t]
    \centering
    \begin{subfigure}{.48\columnwidth}
        \centering
        \includegraphics[width=\columnwidth]{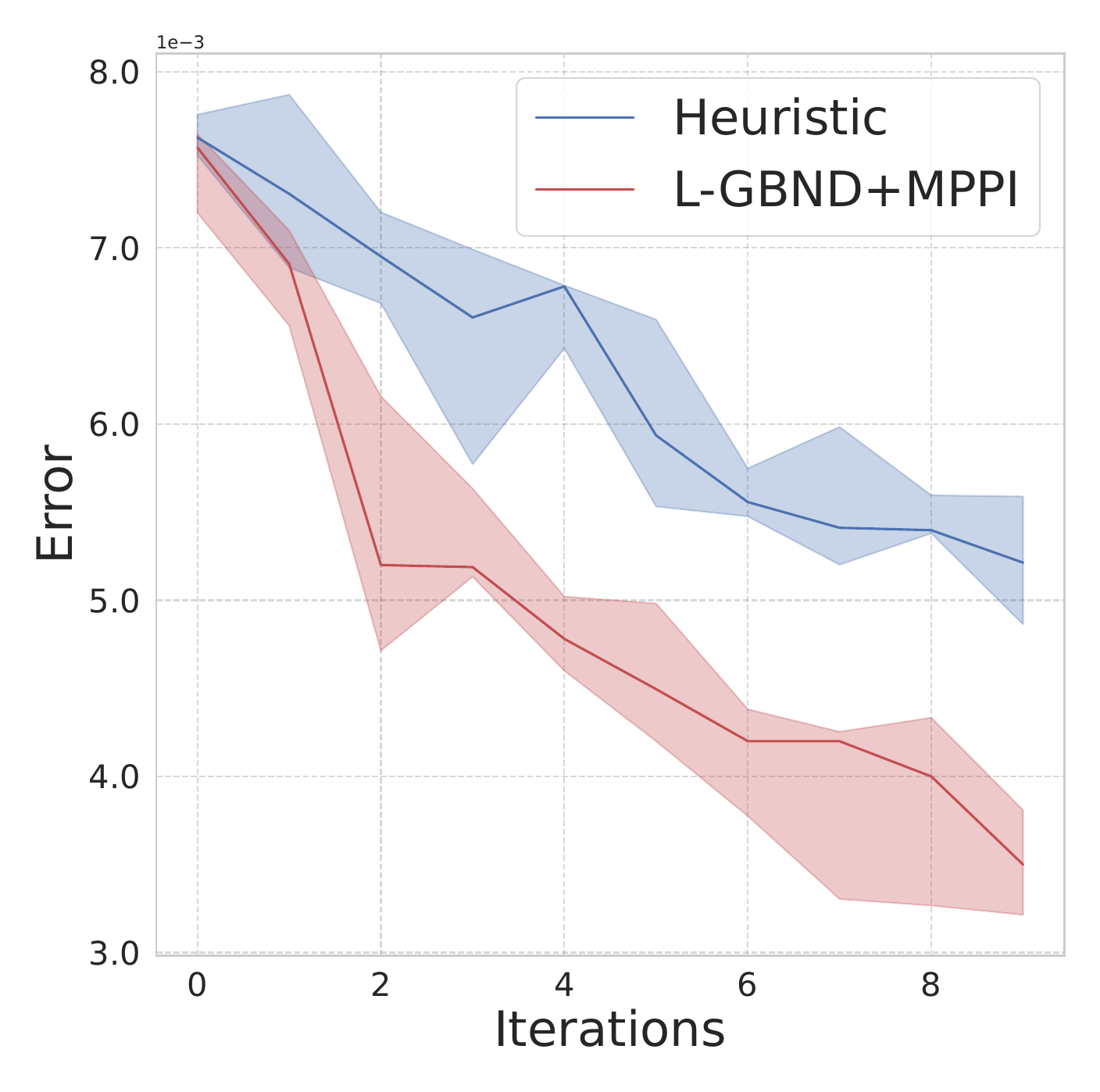}
        \vspace{-18pt}
        \caption{Hole Pebble}
    \end{subfigure}
    \hfill
    \begin{subfigure}{.48\columnwidth}
        \centering
        \includegraphics[width=\columnwidth]{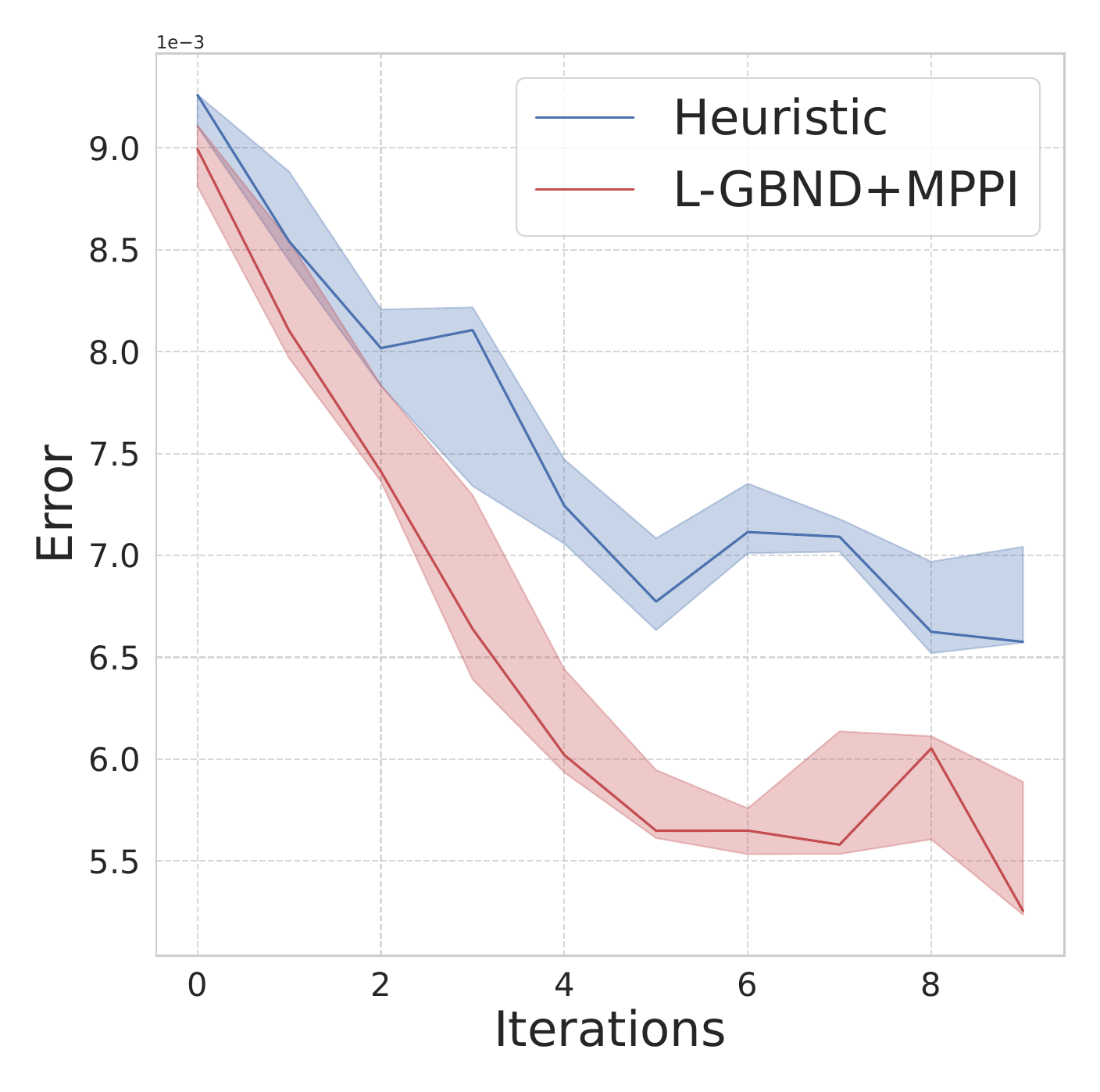}
        \vspace{-18pt}
        \caption{Hole Sand}
    \end{subfigure}
    \hfill
    \begin{subfigure}{.48\columnwidth}
        \centering
        \includegraphics[width=\columnwidth]{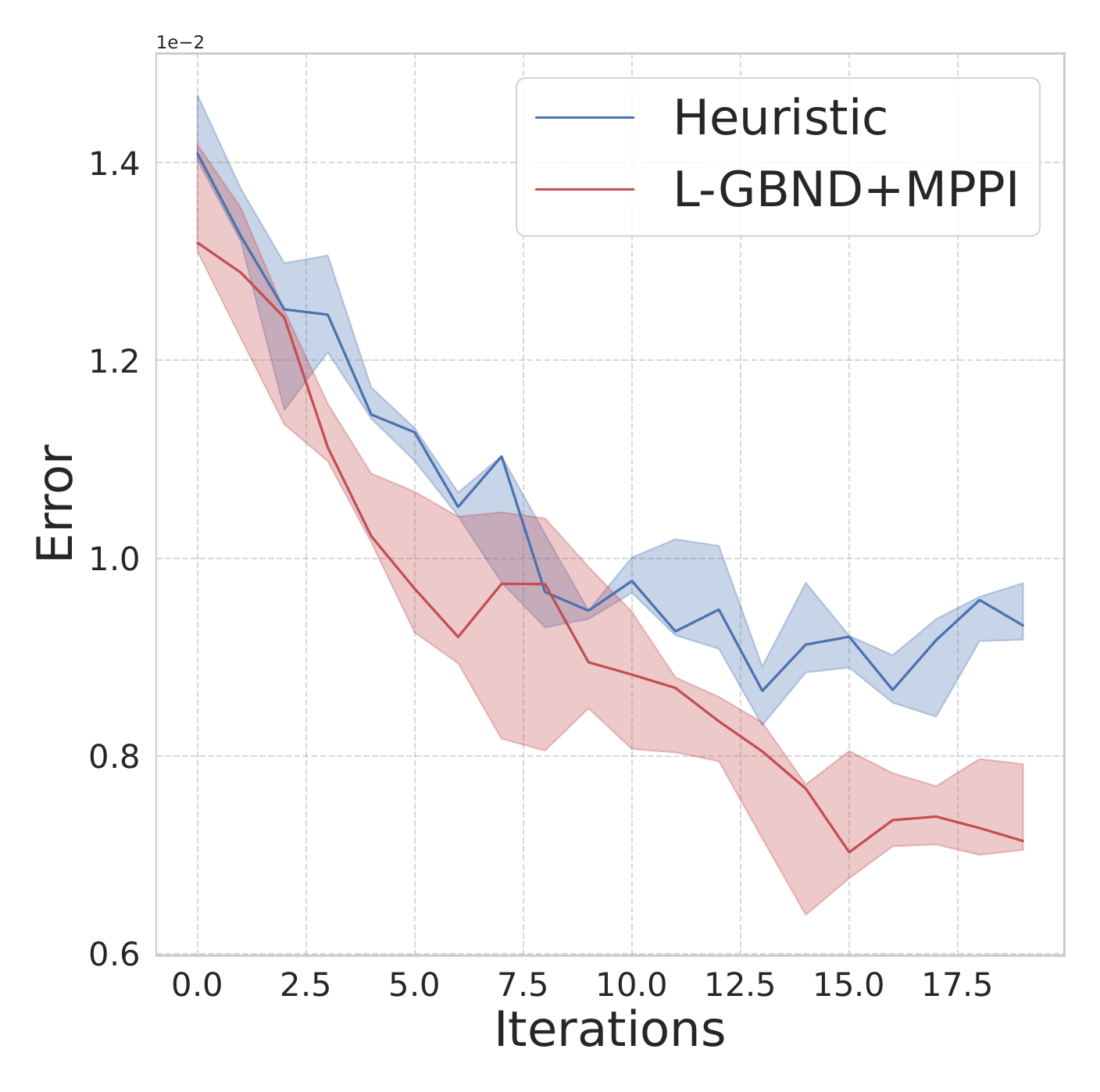}
        \vspace{-18pt}
        \caption{Moat Pebble}
    \end{subfigure}
    \hfill
    \begin{subfigure}{.48\columnwidth}
        \centering
        \includegraphics[width=\columnwidth]{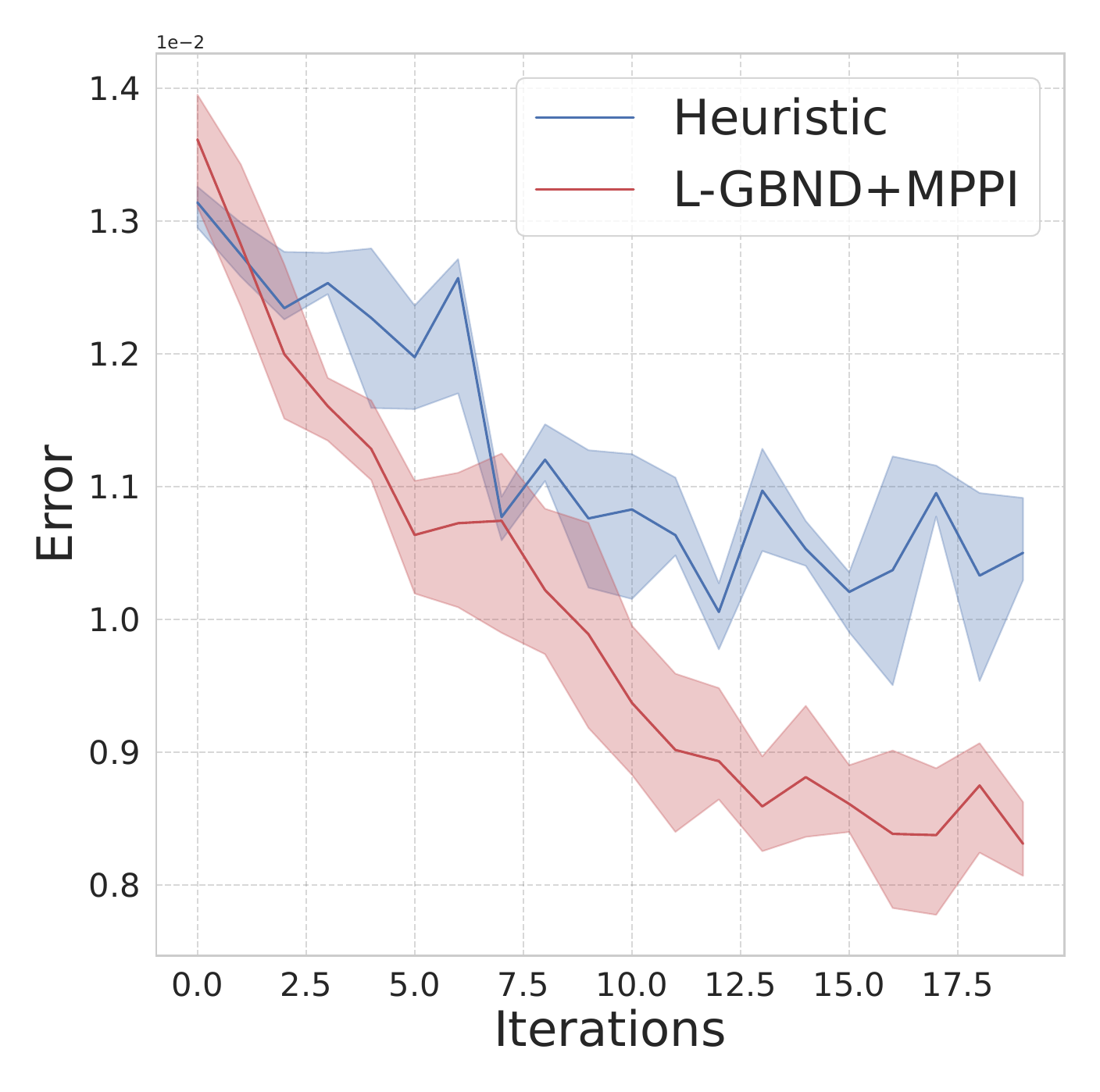}
        \vspace{-18pt}
        \caption{Moat Sand}
    \end{subfigure}
    \caption{Errors between the observed terrain and the target terrain after each PDS trajectory. Interquartile ranges are shaded. 10 trials per scenario.}
    \label{fig:new_task_quantitative}
\end{figure}

For a qualitative comparison, we implement a heuristic baseline that employs dynamic programming to determine the scooping trajectory that maximizes the intersection of the swept volume and the excess material in the current terrain heightmap, similar to the volume maximization approach developed by Yang et al.~\cite{Yang2021}. We also constrain the solver to use axis-aligned trajectories to boost its computational efficiency.  Fig. \ref{fig:new_task_quantitative} illustrates that the heuristic makes rapid initial progress but is unable to shape the terrain's finer details by anticipating settling behavior.

\begin{figure}[t]
    \centering
    \begin{subfigure}{.4\columnwidth}
        \includegraphics[width=\columnwidth]{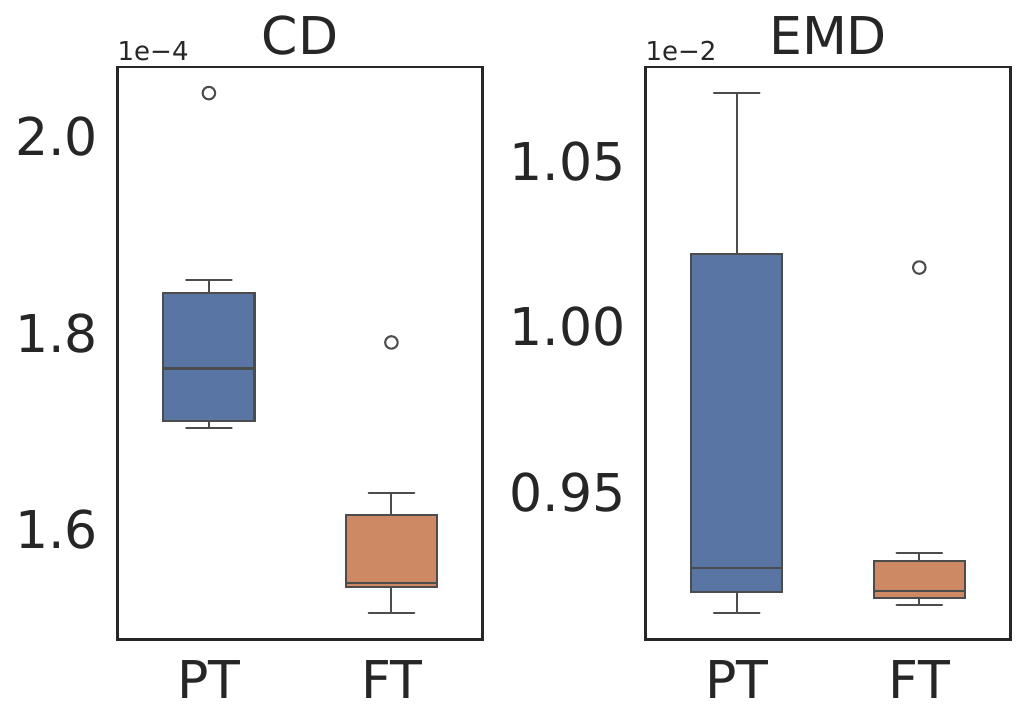}
        \caption{Rock}
    \end{subfigure}
    % \hfill
    \hspace{10pt}
    \begin{subfigure}{.4\columnwidth}
        \includegraphics[width=\columnwidth]{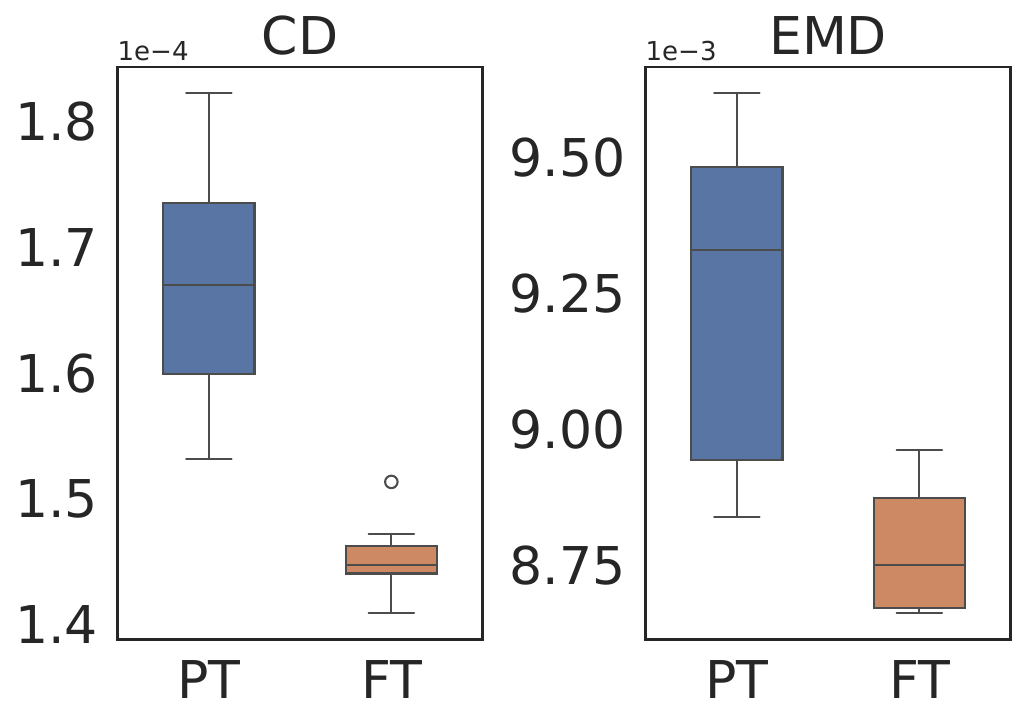}
        \caption{Sand}
    \end{subfigure}
    \vspace{-3pt}
    \caption{Fine-tuning (FT) the pre-trained (PT) L-GBND model on real-world datasets significantly improves the model stability and precision in practical scenarios, as evidenced by a reduced median error and a more compact interquartile range.}
    \label{fig:cd_emd_comparison}
\end{figure}

\section{CONCLUSION AND LIMITATION}

We presented \lgbnd for handling unbounded terrains in graph-based neural dynamics models that simultaneously learns a RoI and particle dynamics.  Novel node features are key to success of this model.  Experiments show that this method achieves orders of magnitude faster prediction and GPU memory usage compared with na\"{i}ve \gbnd models or GPU-based physics simulators.  It can be trained with both simulated and real data, and when used as a model for planning our method can achieve desired target shapes on real-world excavation platforms.

There remains some limitations of our work that we would like to address in future work. First, our RoI proposer does not accurately model long-range terrain effects (e.g., landslides).  Second, our current platform uses an overhead camera to update the terrain, and we would like to extend this to egocentric 3D mapping, in which our model may need to model occluded parts of the terrain.  Third, our models assume terrain homogeneity, and future work should address heterogeneous materials including terrains with embedded obstacles. Lastly, the current dynamics model requires full-parameter fine-tuning when moving to unseen materials, one potential solution is to adapt physics-parameter conditioning~\cite{zhang2024adaptigraph} for online adaptation.

\bibliographystyle{plainnat}
\bibliography{reference}

\end{document}